\documentclass{article} 
\usepackage{iclr2016_conference,times}
\usepackage{hyperref}
\usepackage{url}
\usepackage[utf8]{inputenc}
\usepackage{enumerate}
\usepackage{algorithm}
\usepackage{algorithmic}
\usepackage{amsmath}
\usepackage{amsfonts}
\usepackage{tabularx}
\usepackage{float}
\usepackage{graphicx}
\usepackage{epstopdf}

\usepackage{subfig}
\newcommand*\samethanks[1][\value{footnote}]{\footnotemark[#1]}

\title{Clustering is Efficient for Approximate\\ Maximum Inner Product Search}

\author{Alex~Auvolat\thanks{Equal contribution}\\
École Normale Supérieure, France.\\
\And
Sarath~Chandar\samethanks ~, Pascal~Vincent\samethanks~~\thanks{and CIFAR}\samethanks\\
Université de Montréal, Canada.\\
\AND
Hugo~Larochelle\samethanks\\
Twitter Cortex, USA.,\\
Université de Sherbrooke, Canada \\
\And
~\And~\And~\And
Yoshua~Bengio\samethanks\\
Université de Montréal, Canada.\\
}

\newcommand{\argmin}{\mathrm{argmin}}
\newcommand{\argmax}{\mathrm{argmax}}

\begin{document}

\maketitle

\begin{abstract}
Efficient \textit{Maximum Inner Product Search} (MIPS) is an important task that has a wide applicability in  recommendation systems and classification with a large number of classes. Solutions based on locality-sensitive hashing (LSH) as well as tree-based solutions have been investigated in the recent literature, to perform approximate MIPS in sublinear time. In this paper, we compare these to another extremely simple approach for solving approximate MIPS, based on variants of the $k$-means clustering algorithm. Specifically, we propose to train a spherical $k$-means, after having reduced the MIPS problem to a Maximum Cosine Similarity Search (MCSS). Experiments on two standard recommendation system benchmarks as well as on large vocabulary word embeddings, show that this simple approach yields much higher speedups, for the same retrieval precision, than current state-of-the-art hashing-based and tree-based methods. This simple method also yields more robust retrievals when the query is corrupted by noise.
\end{abstract}

\section{Introduction}
The {\em Maximum Inner Product Search} (MIPS) problem has recently received
increased attention, as it arises naturally in many large scale tasks. In recommendation systems \citep{cf-iptree, xbox}, users and items to be recommended are represented as vectors that are learnt at training time based on the user-item rating matrix. At test time, when the model is deployed for suggesting recommendations, given a user vector, the model will perform a dot product of the user vector with all the item vectors and pick top $K$ items with maximum dot product to recommend. With millions of candidate items to recommend, it is usually not possible to do a full linear search within the available time frame of only few milliseconds. This problem amounts to solving a $K$-MIPS problem. 

Another common instance where the $K$-MIPS problem arises is in extreme classification tasks \citep{wta}, with a huge number of classes.
At inference time, predicting the top-$K$ most likely class labels for a given data point can be cast as a $K$-MIPS problem.
Such extreme (probabilistic) classification problems occur often in Natural Language Processing (NLP) tasks where the classes are words in a predetermined vocabulary. For example in neural probabilistic language models~\citep{Bengio:2003} the probabilities of a next word given the context of the few previous words is computed, in the last layer of the network, as a multiplication of the last hidden layer representation with a very large matrix (an embedding dictionary) that has as many columns as there are words in the vocabulary. Each such column can be seen as corresponding to the embedding of a vocabulary word in the hidden layer space. Thus an inner product is taken between each of these and the hidden representation, to yield an inner product ``score'' for each vocabulary word. Passed through a softmax nonlinearity, these yield the predicted probabilities for all possible words. The ranking of these probability values is unaffected by the softmax layer, so finding the $k$ most probable words is exactly equivalent to finding the ones with the largest inner product scores, i.e.\ solving a $K$-MIPS problem. 
 
 In many cases the retrieved result need not be exact: it may be sufficient to obtain a subset of $k$ vectors whose inner product with the query is very high, and thus highly likely (though not guaranteed) to contain some of the exact $K$-MIPS vectors.
 These examples motivate research on approximate $K$-MIPS algorithms. If we can obtain large speedups over a full linear search without sacrificing too much on precision, it will have a direct impact on such large-scale applications.
 
Formally the $K$-MIPS problem
is stated as follows: given a set $\mathcal{X}=\{x_1, \dots, x_n\}$ of points and a query vector $q$, find
\begin{equation}
\argmax^{(K)}_{i \in \mathcal{X}} ~~ q^\top x_i
\end{equation}
where the $\argmax^{(K)}$ notation corresponds to the set of the indices providing the $K$ maximum values. Such a problem can be solved exactly in linear time by calculating all the $q^\top x_i$ and selecting the $K$ maximum items, but such a method is too costly to be used on large applications where we typically have hundreds of thousands of entries in the set.

All the methods discussed in this article are based on the notion of a {\em candidate set}, i.e.\ a subset of the dataset that they return, and on which we will do an exact $K$-MIPS, making its computation much faster. There is no guarantee that the candidate set contains the target elements, therefore these methods solve {\em approximate} $K$-MIPS.
Better algorithms will provide us with candidate sets that are both smaller and have larger intersections with the actual $K$ maximum inner product vectors.

MIPS is related to nearest neighbor search (NNS), and to maximum similarity search. But it is considered a harder problem because the inner product neither satisfies the triangular inequality as distances usually do, nor does it satisfy a basic property of similarity functions, namely that the similarity of an entry with itself is at least as large as its similarity with anything else: for a vector $x$, there is no guarantee that $x^Tx \geq x^Ty$ for all $y$. Thus we cannot \emph{directly} apply efficient nearest neighbor search or maximum similarity search algorithms to the MIPS problem. 

Given a set $\mathcal{X}=\{x_1, \dots, x_n\}$ of points and a query vector $q$, the $K$-NNS problem with Euclidean distance 
is defined as:
\begin{equation}
\argmin^{(K)}_{i \in \mathcal{X}} ~~ ||q - x_i||_2^2 = \argmax^{(K)}_{i \in \mathcal{X}} ~~ q^T x_i - \frac{||x_i||_2^2}{2} 
\end{equation}
and the maximum cosine similarity problem ($K$-MCSS) is defined as:
\begin{equation}
\argmax^{(K)}_{i \in \mathcal{X}} ~~ \frac{q^T x_i}{||q||~~ ||x_i||} = \argmax^{(K)}_{i \in \mathcal{X}} ~~ \frac{q^T x_i}{||x_i||}  
\end{equation}
$K$-NNS and $K$-MCSS are different problems than $K$-MIPS, but it is easy to see that all three become equivalent provided all data vectors $x_i$ have the same Euclidean norm. 
Several approaches to MIPS make use of this observation and first transform a MIPS problem into a NNS or MCSS problem.

In this paper, we propose and empirically investigate a very simple approach for the approximate $K$-MIPS problem. It consists in first reducing the problem to an approximate $K$-MCSS problem (as has been previously done in \citep{shrivastava2014improved} ) on top of which we perform a spherical $k$-means clustering. The few clusters whose centers best match the query yield the candidate set. 


The rest of the paper is organized as follows: In section 2, we review previously proposed approaches for MIPS. Section 3 describes our proposed simple solution \textit{$k$-means MIPS} in more details and section 4 discusses ways to further improve the performance by using a hierarchical $k$-means version. In section 5, we empirically compare our methods to the state-of-the-art in tree-based and hashing-based approaches, on two standard collaborative filtering benchmarks and on a larger word embedding datasets. Section 6 concludes the paper with discussion on future work.

\section{Related Work}
There are two common types of solution for MIPS in the literature: tree-based methods and hashing-based methods. Tree-based methods are data dependent (i.e. first trained to adapt to the specific data set) while hash-based methods are mostly data independent.

\textbf{Tree-based approaches:} The Maximum Inner Product Search problem was first formalized in \citep{Ram:2012}. \cite{Ram:2012} provided a tree-based solution for the problem. Specifically, they constructed a ball tree with vectors in the database and bounded the maximum inner product with a ball. Their novel analytical upper bound for maximum inner product of a given point with points in a ball made it possible to design a branch and bound algorithm to solve MIPS using the constructed ball tree. \cite{Ram:2012} also proposes a dual-tree based search using cone trees when you have a batch of queries. One issue with this ball-tree based approach (IP-Tree) is that it partitions the set of data points based on the Euclidean distance, while the problem hasn't effectively been converted to NNS. In contrast, PCA-Tree \citep{xbox}, the current state-of-the-art tree-based approach to MIPS, first converts MIPS to NNS by appending an additional component to the vector that ensures that all vectors are of constant norm. This is followed by PCA and by a balanced kd-tree style tree construction.

\textbf{Hashing based approaches:}
\cite{Shrivastava014} is the first work to propose an explicit Asymmetric Locality Sensitive Hashing (ALSH) construction to perform MIPS. They converted MIPS to NNS and used the L2-LSH algorithm \citep{Datar}. Subsequently, \cite{shrivastava2014improved} proposed another construction to convert MIPS to MCSS and used the Signed Random Projection (SRP) hashing method. Both works were based on the assumption that a symmetric-LSH family does not exist for MIPS problem. Later, \cite{symmetric} showed an explicit construction of a symmetric-LSH algorithm for MIPS which had better performance than the previous ALSH algorithms. Finally, \cite{wta} propose to use Winner-Take-All hashing to pick top-$K$ classes to consider during training and inference in large classification problems. 

\textbf{Hierarchical softmax:}
A notable approach to address the problem of scaling classifiers to a huge number of classes is the hierarchical softmax~\citep{Morin+al-2005}. It is based on prior clustering of the words into a binary, or more generally $n$-ary tree that serves as a fixed structure for the learning process of the model. The complexity of training is reduced from $O(n)$ to $O(\log n)$. Due to its clustering and tree structure, it resembles the MIPS techniques we explore in this paper. However, the approaches differ at a fundamental level. Hierarchical softmax defines the probability of a leaf node as the product of all the probabilities computed by all the intermediate softmaxes on the way to that leaf node. 
By contrast, an approximate MIPS search imposes no such constraining structure on the probabilistic model, and is better though as efficiently searching for top winners of what amounts to a large ordinary flat softmax.


\section{$k$-means clustering for approximate MIPS}
In this section, we propose a simple $k$-means clustering based solution for approximate MIPS. 

\subsection{MIPS to MCSS}
We follow the previous work by ~\cite{shrivastava2014improved} for reducing the MIPS problem to the MCSS problem by ingeniously rescaling the vectors and adding new components, making the norms of all the vectors approximately the same. 
Let $\mathcal{X} = \{x_1, \dots, x_n\}$ be our dataset. Let $U < 1$ and $m \in \mathbb{N}^{*}$ be parameters of the algorithm. The first step is to scale all the vectors in our dataset by the same factor such that $\max_i ||x_i||_2 = U$. We then apply two mappings $P$ and $Q$, one on the data points and another on the query vector. These two mappings simply concatenate $m$ new components to the vectors
making the norms of the data points all roughly the same. The mappings are defined as follows:
\begin{eqnarray}
    P(x) & = & [x, 1/2 - ||x||_2^2,  1/2 - ||x||_2^4, \dots, 1/2 - ||x||_2^{2^m}] \\
    Q(x) & = & [x, 0, 0, \dots, 0]
\end{eqnarray}
As shown in~\cite{shrivastava2014improved}, mapping $P$ brings all the vectors to roughly the same norm: we have $||P(x_i)||_2^2 = m/4 + ||x_i||_2^{2^{m+1}}$,
with the last term vanishing as $m \to +\infty$, since $||x_i||_2 \leq U < 1$. We thus have the following approximation of MIPS by MCSS
for any query vector $q$,
\begin{eqnarray}
    \argmax^{(K)}_i q^\top x_i & \simeq
         & \argmax^{(K)}_i \frac{Q(q)^\top P(x_i)}{||Q(q)||_2 \cdot ||P(x_i)||_2}
\end{eqnarray}

\subsection{MCSS using Spherical $k$-means}
Assuming all data points $x_1, \dots, x_n$ have been transformed as $x_j \leftarrow P(x_j)$ so as to be scaled to a norm of approximately 1,
then the {\em spherical $k$-means\footnote{Note that we use $K$ to refer to the number of top-$K$ items to retrieve in search and $k$ for the number of clusters in $k$-means. These two quantities are otherwise not the same.}} algorithm ~\citep{zhong2005efficient}
can efficiently be used to do approximate MCSS. Algorithm~\ref{sphkm} is a formal specification of the spherical $k$-means algorithm, where we denote by $c_i$ the centroid of cluster $i$ ($i \in \{1, \dots, K\}$) and $a_j$ the index of the cluster assigned to each point $x_j$.
    \begin{algorithm}[H]
    \caption{Spherical $k$-means}
    \label{sphkm}
    \begin{algorithmic}
    \STATE $a_j \gets \mathrm{rand}(k)$
    \WHILE {$c_i$ or $a_j$ changed at previous step}
        \STATE $c_i \gets \frac{\sum_{j | a_j = i} x_j}{|| \sum_{j | a_j = i} x_j ||}$
        \STATE $a_j \gets \argmax_{i \in \{1, \dots, k\}} x_j^\top c_i$
    \ENDWHILE
    \end{algorithmic}
    \end{algorithm}
The difference between standard $k$-means clustering and spherical $k$-means is that in the spherical variant, the data points are clustered not according to their position in the Euclidian space, but according to their direction. 

To find the one vector that has maximum cosine similarity to query point $q$ in a dataset clustered by this method, we first find the cluster whose centroid has the best cosine similarity with the query vector -- i.e.\ the $i$ such that $q^\top c_i$ is maximal -- and consider all the points belonging to that cluster as the candidate set. We then simply take $\argmax_{j | a_j = i}\; q^\top x_j$ as an approximation for
our maximum cosine similarity vector. This method can be extended for finding the $k$ maximum cosine similarity vectors: we compute the cosine similarity between the query and all the vectors of the candidate set and take the $k$ best matches. 

One issue with constructing a candidate set from a single cluster is that the quality of the set will be poor for points close to the boundaries between clusters. To alleviate this problem, we can increase the size of candidate sets by constructing them instead from the top-$p$ best matching clusters to construct our candidate set. 

We note that other approximate search methods exploit similar ideas. For example, \cite{xbox} proposes a so-called {\it neighborhood boosting} method for PCA-Tree, by considering the path to each leaf as a binary vector (based on decision to go left or right) and given a target leaf, consider all other leaves which are one hamming distance away.


\section{Hierarchical $k$-means for faster and more precise search}

While using a single-level clustering of the data points might yield a sufficiently fast search procedure for moderately large databases, it can be insufficient for much larger collections.

Indeed, if we have $n$ points, by clustering our dataset into $\sqrt{n}$ clusters so that each cluster contains approximately $\sqrt{n}$ points, we reduce the complexity of the search from $O(n)$ to roughly $O\left(\sqrt{n}\right)$. If we use the single closest cluster as a candidate set, then the candidate set size is of the order of $\sqrt{n}$. But as mentioned earlier, we will typically want to consider the two or three closest clusters as a candidate set, in order to limit problems arising from the query points close to the boundary between clusters or when doing approximate $K$-MIPS with $K$ fairly big (for example 100). A consequence of increasing candidate sets this way is that they can quickly grow wastefully big, containing many unwanted items. To restrict the candidate sets to a smaller count of better targeted items, we would need to have smaller clusters, but then the search for the best matching clusters becomes the most expensive part. To address this situation, we propose an approach where we cluster our dataset into many small clusters, and then cluster the small clusters into bigger clusters, and so on any number of times. Our approach is thus a bottom-up clustering approach.

For example, we can cluster our datasets in $n^{2/3}$ first-level, small clusters, and then cluster the centroids of the first-level clusters into $n^{1/3}$ second-level clusters, making our data structure a two-layer hierarchical clustering.
This approach can be generalized to as many levels of clustering as necessary.

\begin{figure}[htb]
\centering
\begin{minipage}{.8\linewidth}
\centering
    \includegraphics[width=0.8\textwidth]{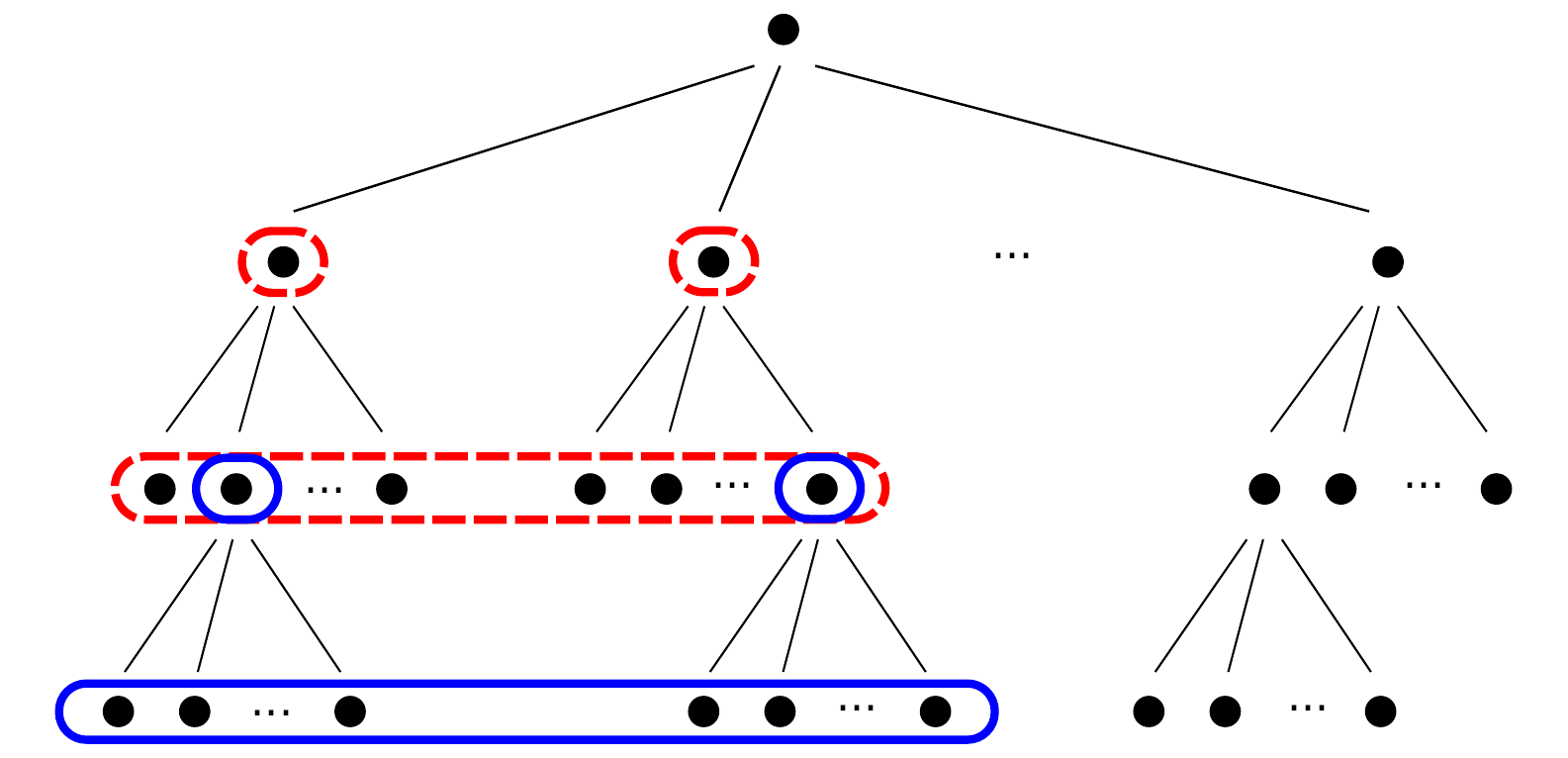}
    \caption{Walk down a hierarchical clustering tree: at each level we have a candidate set
    for the next level. In the first level, the dashed red boxed represent the $p$ best matches, which gives
        us a candidate set for the second level, etc.}
    \label{hierwalk}
\end{minipage}
\end{figure}

To search for the small clusters that best match the query point and will constitute a good candidate set,
we go down the hierarchy keeping at each level only the $p$ best matching clusters.
This process is illustrated in Figure~\ref{hierwalk}.
Since at all levels the clusters are of much smaller size, we can take much
larger values for $p$, for example $p=8$ or $p=16$.

Formally, if we have $L$ levels of clustering, let $I_l$ be a set of indices
for the clusters at level $l\in\{0,\dots,L\}$.
Let $c^{(l)}_i,i\in I_l$ be the centroids of the clusters at level $l$,
with $\{ c^{(L)}_i\}$ conveniently defined as being the data points themselves,
and let $a^{(l)}_i \in I_{l-1}, i\in I_l$ be the assignment of the centroids $c^{(l)}_i$ to the
clusters of layer $l-1$. The candidate set is found using the method described in Algorithm~\ref{treewalk}.
Our candidate set is the set $C_L$ obtained at the end of the algorithm.
\begin{algorithm}
\caption{Search in hierarchical spherical $k$-means}
\label{treewalk}
\begin{algorithmic}
\STATE $C_0 = I_0$
\FOR {$l = 0, \dots, L-1$}
    \STATE $A_l = \argmax^{(p)}_{i \in C_l} q^\top c^{(l)}_i$
    \STATE $C_{l+1} = \left\{ i | a^{(l+1)}_i \in A_l \right\}$
\ENDFOR
\STATE \textbf{return} $C_L$
\end{algorithmic}
\end{algorithm}
In our approach, we do a bottom-up clustering, i.e.\ we first cluster the dataset into small clusters,
then we cluster the
small cluster into bigger clusters, and so on until we get to the top level which is only one cluster.
Other approaches have been suggested such as in~\citep{MnihHinton2009},
where the method employed is a top-down clustering strategy
where at each level the points assigned to the current cluster are divided in smaller clusters. The approach
of~\citep{MnihHinton2009} also addresses the problem that using a single lowest-level cluster as a candidate set
is an inaccurate solution by having the data points be in multiple clusters. We use an alternative solution that
consists in exploring several branches of the clustering hierarchy in parallel.


\section{Experiments}
In this section, we will evaluate the proposed algorithm for approximate MIPS. Specifically, we analyze the following characteristics: speedup, compared to the exact full linear search, of retrieving top-$K$ items with largest inner product, and robustness of retrieved results to noise in the query.

\subsection{Datasets}
We have used 2 collaborative filtering datasets and 1 word embedding dataset, which are descibed below:

\textbf{Movielens-10M:} A collaborative filtering dataset with 10,677 movies (items) and 69,888 users. Given the user-item matrix $Z$, we follow the pureSVD procedure described in \citep{Cremonesi:2010} to generate user and movie vectors. Specifically, we subtracted the average rating of each user from his individual ratings and considered unobserved entries as zeros. Then we compute an SVD approximation of $Z$ with its top 150 singular components, $Z \simeq W\Sigma R^T$. Each row in $W\Sigma$ is used as the vector representation of the user and each row in $R$ is the vector representation of the movie. We construct a database of all 10,677 movies and consider 60,000 randomly selected users as queries. 

\textbf{Netflix:} Another standard collaborative filtering dataset with 17,770 movies (items) and 480,189 users. We follow the same procedure as described for movielens but construct 300 dimensional vector representations, as is standard in the literature \citep{symmetric}. We consider 60,000 randomly selected users as queries. 

\textbf{Word2vec embeddings:} We use the 300-dimensional word2vec embeddings released by \citet{NIPS2013_5021}. We construct a database composed of the first 100,000 word embedding vectors. We consider two types of queries: 2,000 randomly selected word vectors from that database, and 2,000 randomly selected word vectors from the database corrupted with Gaussian noise. This acts as a test bench to evaluate the performance of different algorithms based on the characteristics of the queries.

\subsection{Baselines}
We consider the following baselines to compare with.

\textbf{PCA-Tree:} PCA-Tree \citep{xbox} is the state-of-the-art tree-based method which was shown to be superior to IP-Tree \citep{cf-iptree}. This method first converts MIPS to NNS by appending an additional component to the vectors to make them of constant norm. Then the principal directions are learnt and the data is projected using these principal directions. Finally, a balanced tree is constructed using as splitting criteria at each level the median of component values along the corresponding principal direction. Each level uses a different principal direction, in decreasing order of variance.

\textbf{SRP-Hash:} This is the signed random projection hashing method for MIPS proposed in \cite{shrivastava2014improved}. SRP-Hash converts MIPS to MCSS by vector augmentation. We consider $n$ hash functions and each hash function considers $p$ random projections of the vector to compute the hash.

\textbf{WTA-Hash:} Winner Takes All hashing \citep{wta} is another hashing-based baseline which also converts MIPS to MCSS by vector augmentation. We consider $n$ hash functions and each hash function does $p$ different random permutations of the vector. Then the prefix constituted by the first $k$ elements of each permuted vector is used to construct the hash for the vector.

\subsection{Speedup Results}
In these first experiments, we consider the two collaborative filtering tasks and evaluate the speedup provided by the different approximate $K$-MIPS algorithms (for $K \in \{1,10,100\}$) compared to the exact full search. 
Note that this section does not include the hierarchical version of $k$-means in the experiments, as the databases were small enough (less than 20,000) for flat $k$-means to perform well.

Specifically, speedup is defined as
\begin{equation}
\rm{speedup}_{A_0}(A) = \frac{ \rm{Time~taken~by~Algorithm~A_0} } { \rm{Time~taken~by~Algorithm~A} }
\end{equation}
where $A_0$ is the exact linear search algorithm that consists in computing the inner product with all training items. Because we want to compare the preformance of algorithms, rather than of specifically optimized implementations, we approximate the time with the number of dot product operations computed by the algorithm\footnote{For example, $k$-means algorithm was run using GPU while PCA-Tree was run using CPU.}. In other words, our unit of time is the time taken by a dot product.
All algorithms return a set of candidates for which we do exact linear seacrh. This induces a number of dot products at least as large as the size of the identified candidate set.
In addition to the candidate set size, the following operations count towards the count of dot products:

\textbf{$k$-means:} dot products done with all cluster centroids involved in finding the top-$p$ clusters of the (hierarchical) search.

\textbf{PCA-Tree:} dot product done to project the query to the PCA space. Note that if the tree is of depth $d$, then we need to do $d$ dot products to project the query.

\textbf{SRP-Hash:} total number of random projections of the data (each random projection is considered a single dot product). If we have $n$ hashes with $p$ random projections each, then the cost is $p * n$.

\textbf{WTA-Hash:} a full random permutation of the vector involves the same number of query element access operations as a single dot product. However, we consider only $k$ prefixes in the permutations, which means we only need to do a fraction of dot product. 
While a dot product involves accessing all $d$ components of the vector, each permutation in WTA-Hash only needs to access $k$ elements of the vector. So we consider its cost to be a fraction $k/d$ of the cost of a dot product. 
Specifically, if we have $n$ hash functions each with $p$ random permutations and consider prefixes of length $k$, then the total cost would be $n * p * k/d$ where $d$ is the dimension of the vector.

\begin{figure}[ht]
\subfloat[]{\includegraphics[width = 1.8in]{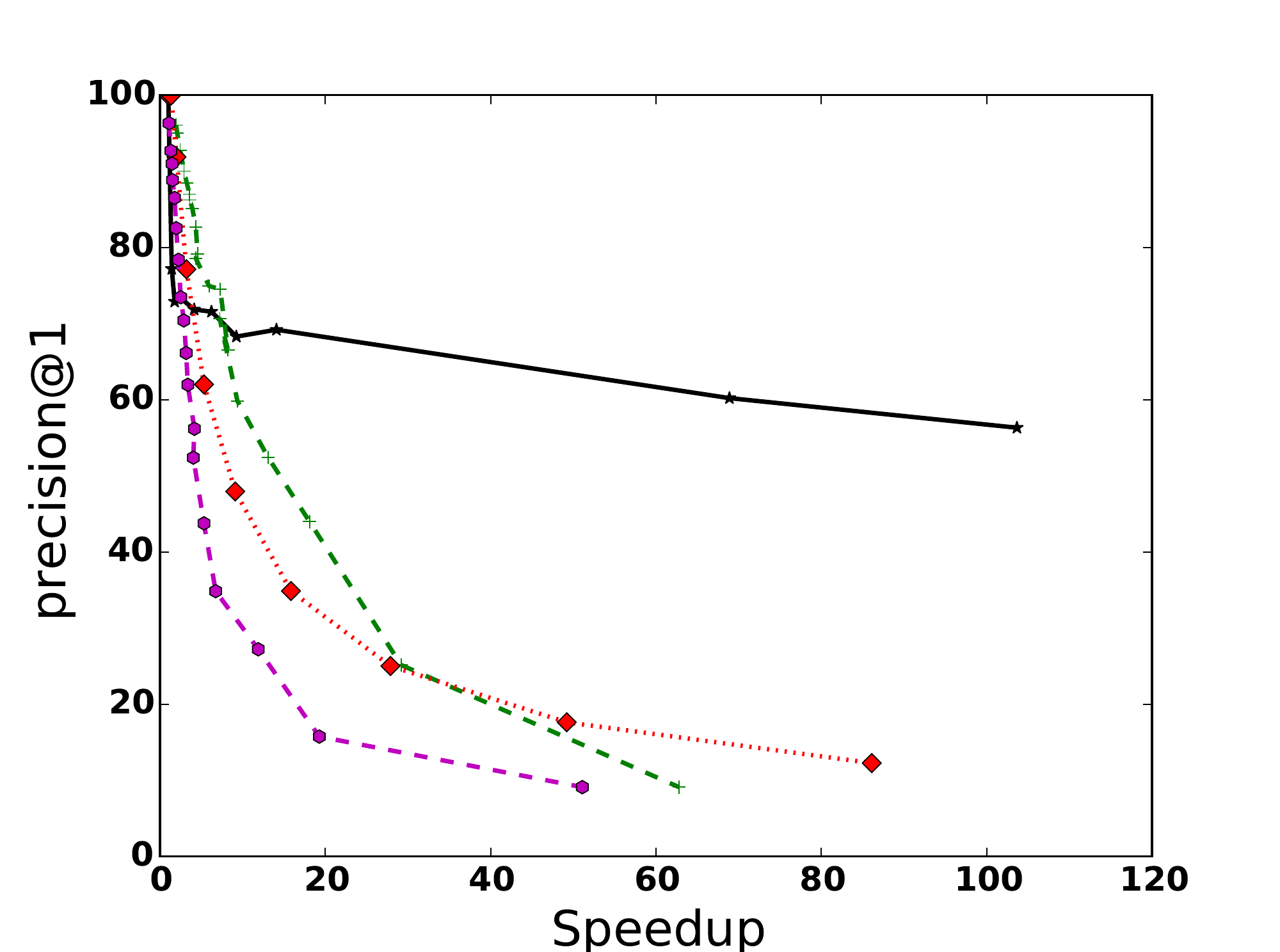}} 
\subfloat[]{\includegraphics[width = 1.8in]{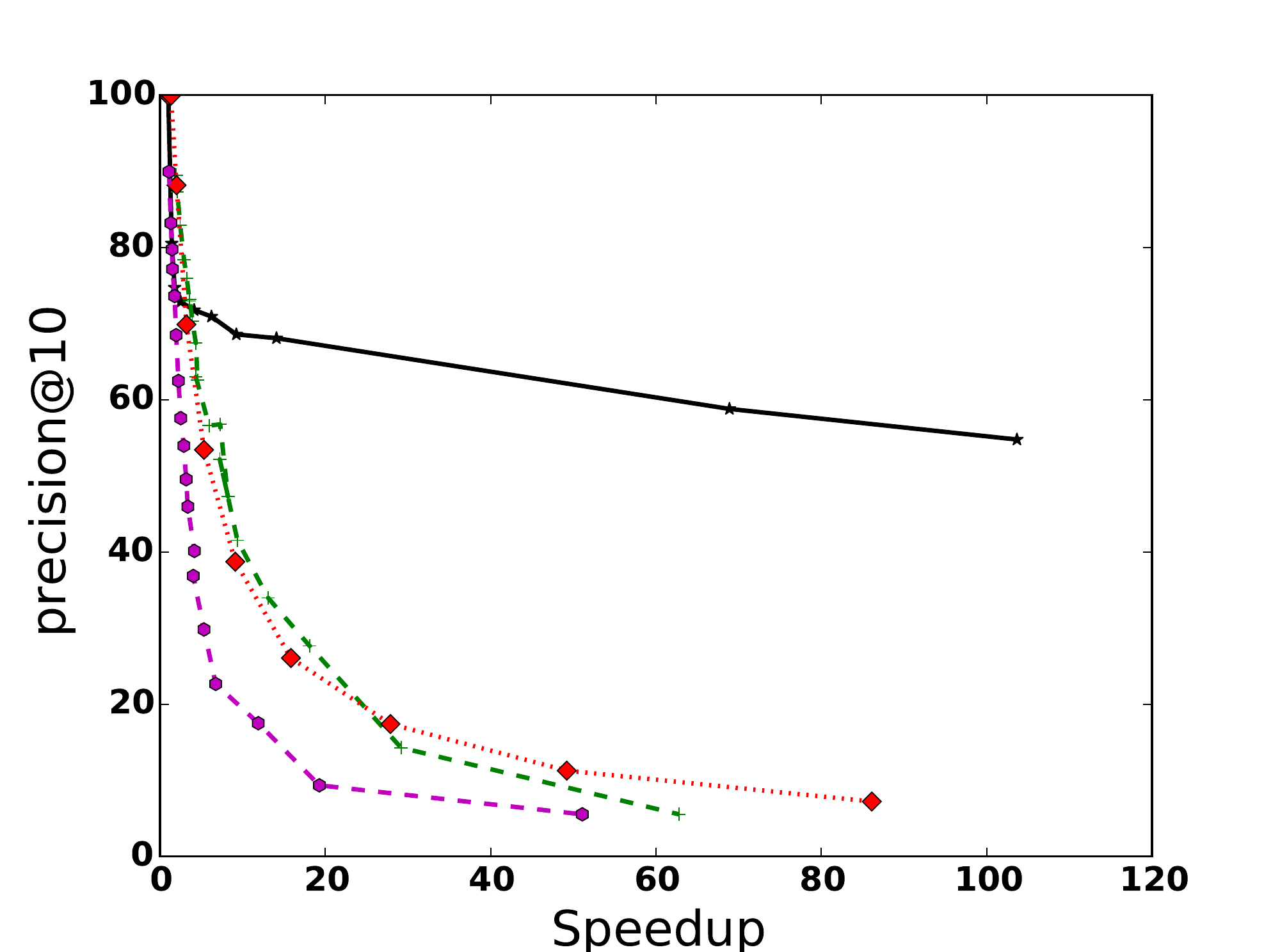}}   
\subfloat[]{\includegraphics[width = 1.8in]{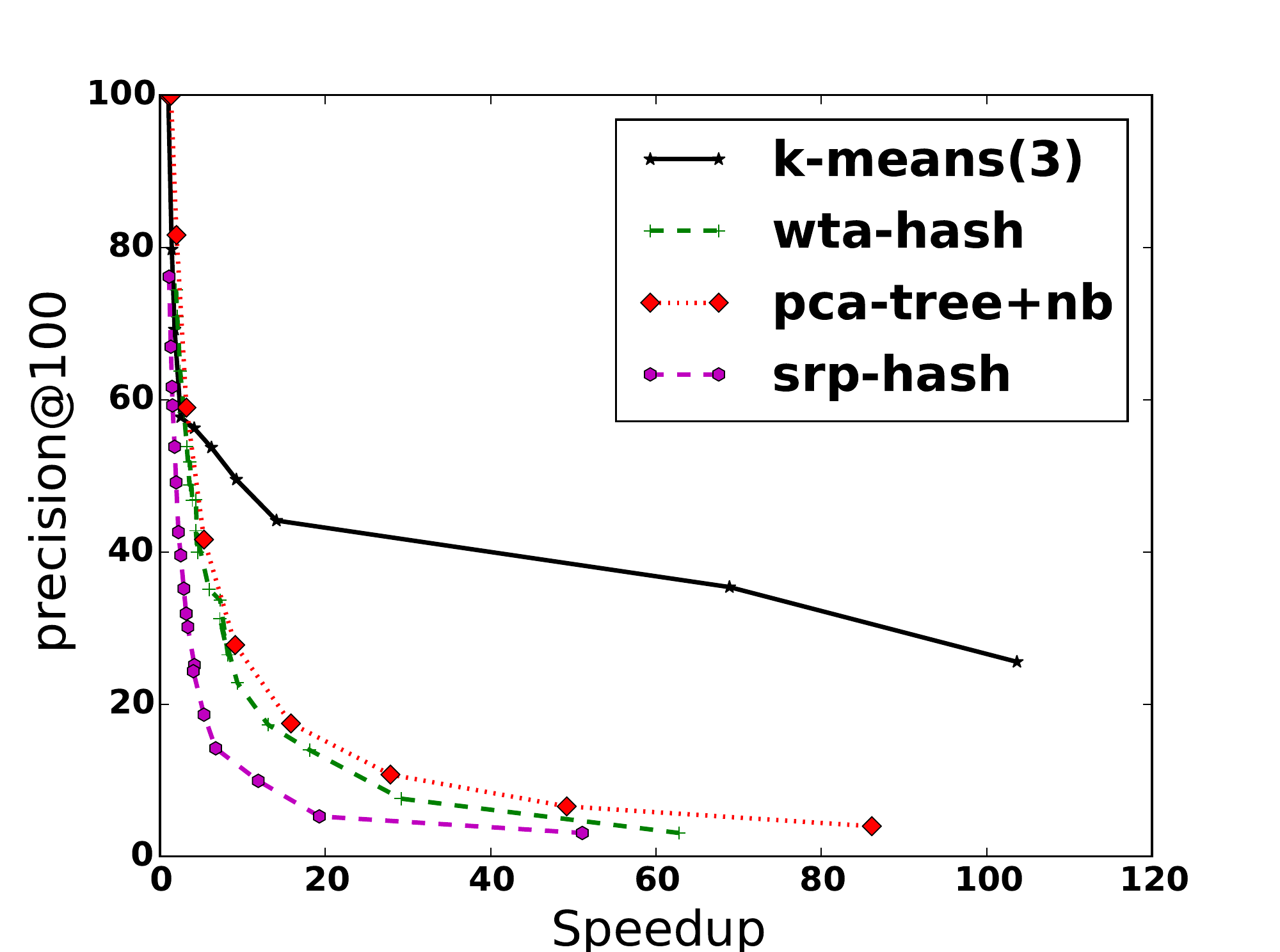}}\\
\subfloat[]{\includegraphics[width = 1.8in]{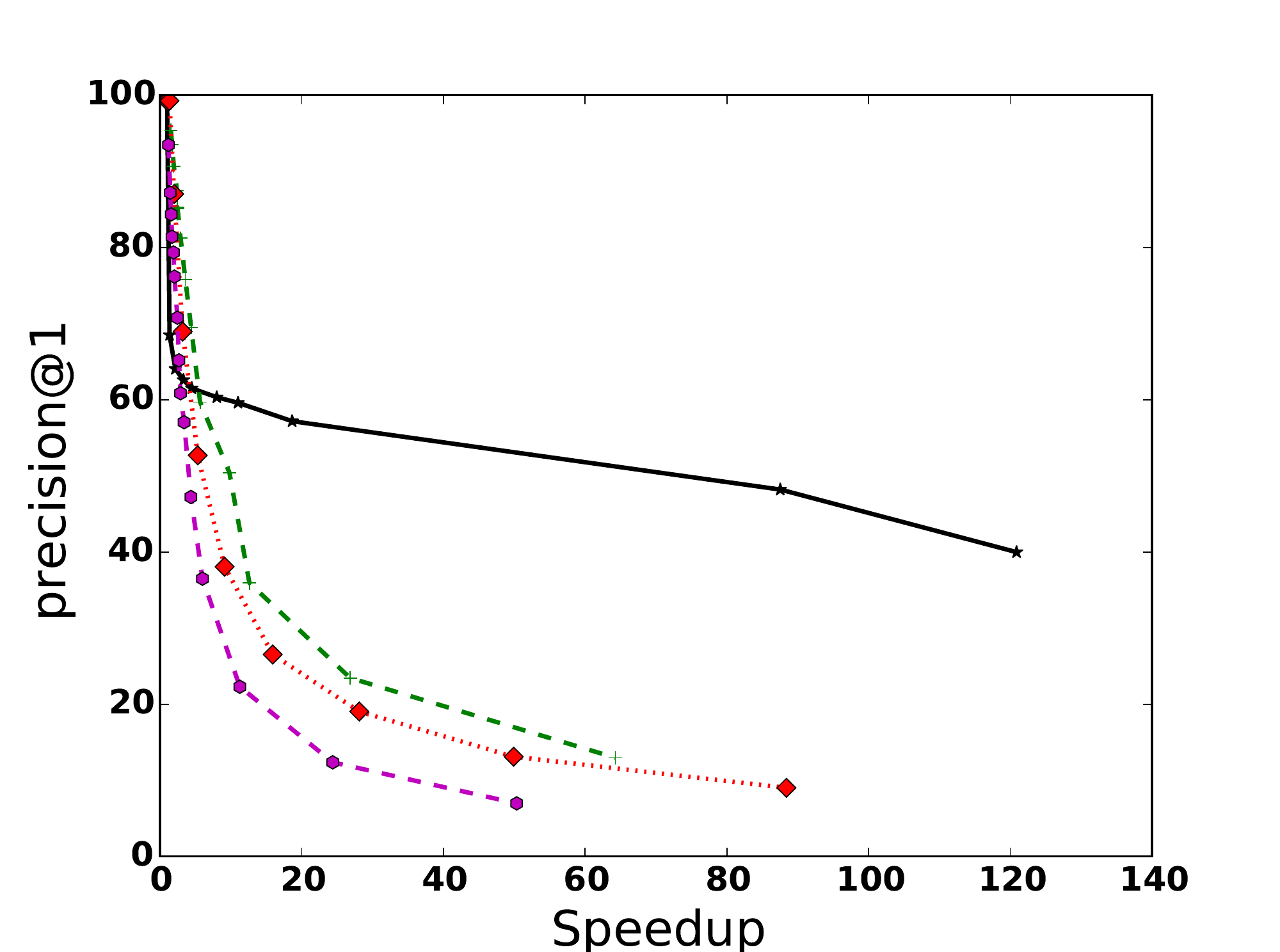}}
\subfloat[]{\includegraphics[width = 1.8in]{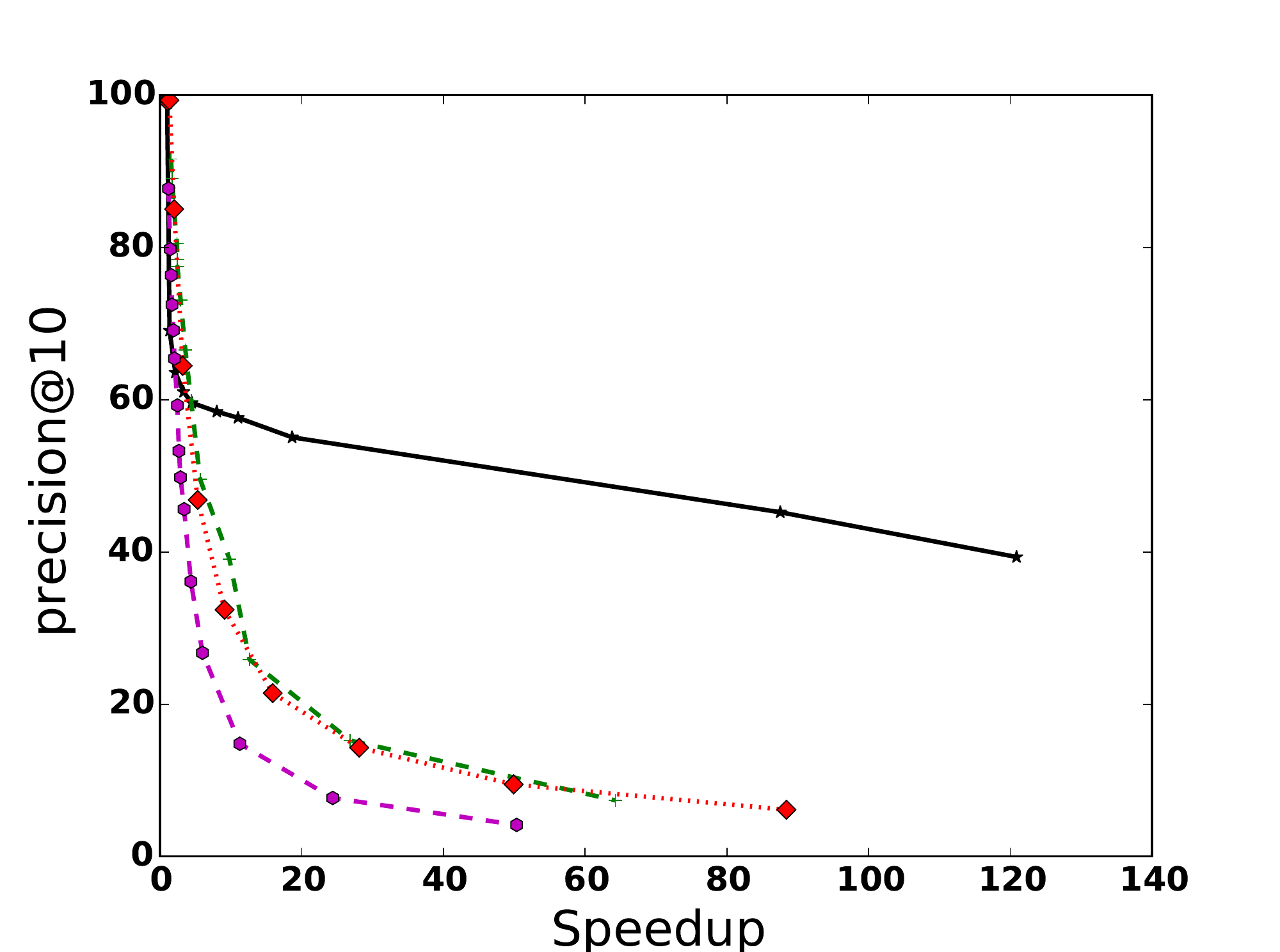}}
\subfloat[]{\includegraphics[width = 1.8in]{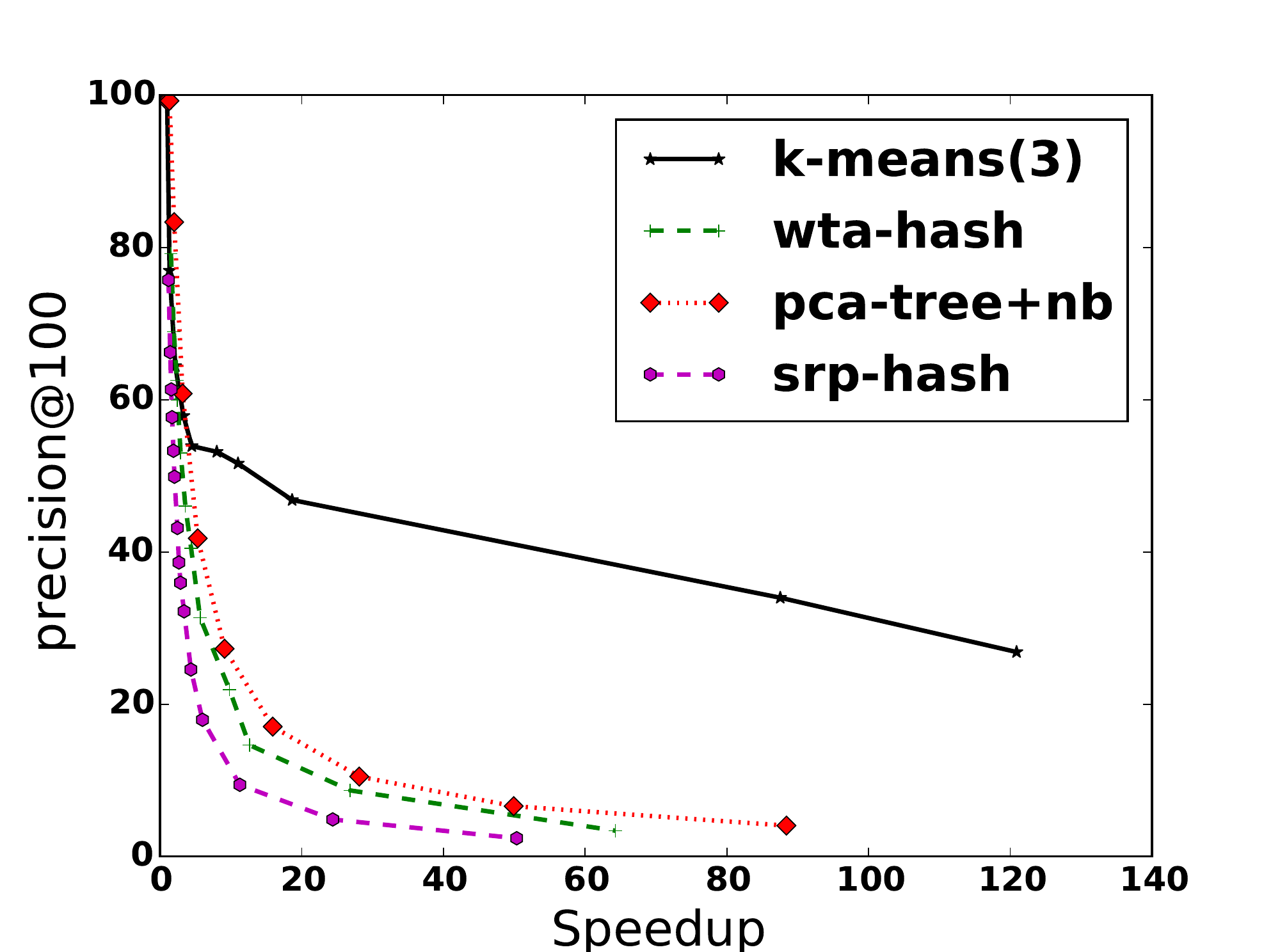}}\\
\caption{Speedup results in collaborative filtering. (a-c) correspond to precision in top 1,10,100 MIPS on Movielens-10M dataset, while (d-f) correspond to precision in top 1,10,100 MIPS on Netflix dataset respectively. $k$-means(3) means $k$-means algorithm that considers top 3 clusters as candidate set.}
\label{ef1}
\end{figure}

Let us call {\it true top-$K$} the actual $K$ elements from the database that have the largest inner products with the query. Let us call {\it retrieved top-$K$} the $K$ elements, among the candidate set retrieved by a specific approximate MIPS, that have the largest inner products with the query.
We define precision for $K$-MIPS as the number of elements in the intersection of true top-$K$ and retrived top-$K$ vectors, divided by $K$.
\begin{equation}
\mathrm{precision~at~}K = \frac{ | \mathrm{retrieved~top~}K \cap \mathrm{true~top~}K | } {K}
\end{equation}
We varied hyper-parameters of each algorithm ($k$ in $k$-means, depth in PCA-Tree, number of hash functions in SRP-Hash and WTA-Hash), and computed the precision and speedup in each case. 
Resulting precision v.s. speedup curves obtained for the Movielens-10M and Netflix datasets are reported in Figure~\ref{ef1}. We make the following observations from these results:
\begin{itemize}
\item Hashing-based methods perform better with lower speedups. But their performance decrease rapidly after 10x speedup.
\item PCA-Tree performs better than SRP-Hash.
\item WTA-Hash performs better than PCA-Tree with lower speedups. However, their performance degrades faster as the speedup increases and PCA-Tree outperforms WTA-Hash with higer speedups.
\item $k$-means is a clear winner as the speed up increases. Also, performance of $k$-means degrades very slowly with increase in speedup as compared to rapid decrease in performance of other algorithms.
\end{itemize}

\subsection{Neighborhood preserving and Robustness Results}
\begin{figure}[ht]
\subfloat[]{\includegraphics[width = 1.8in]{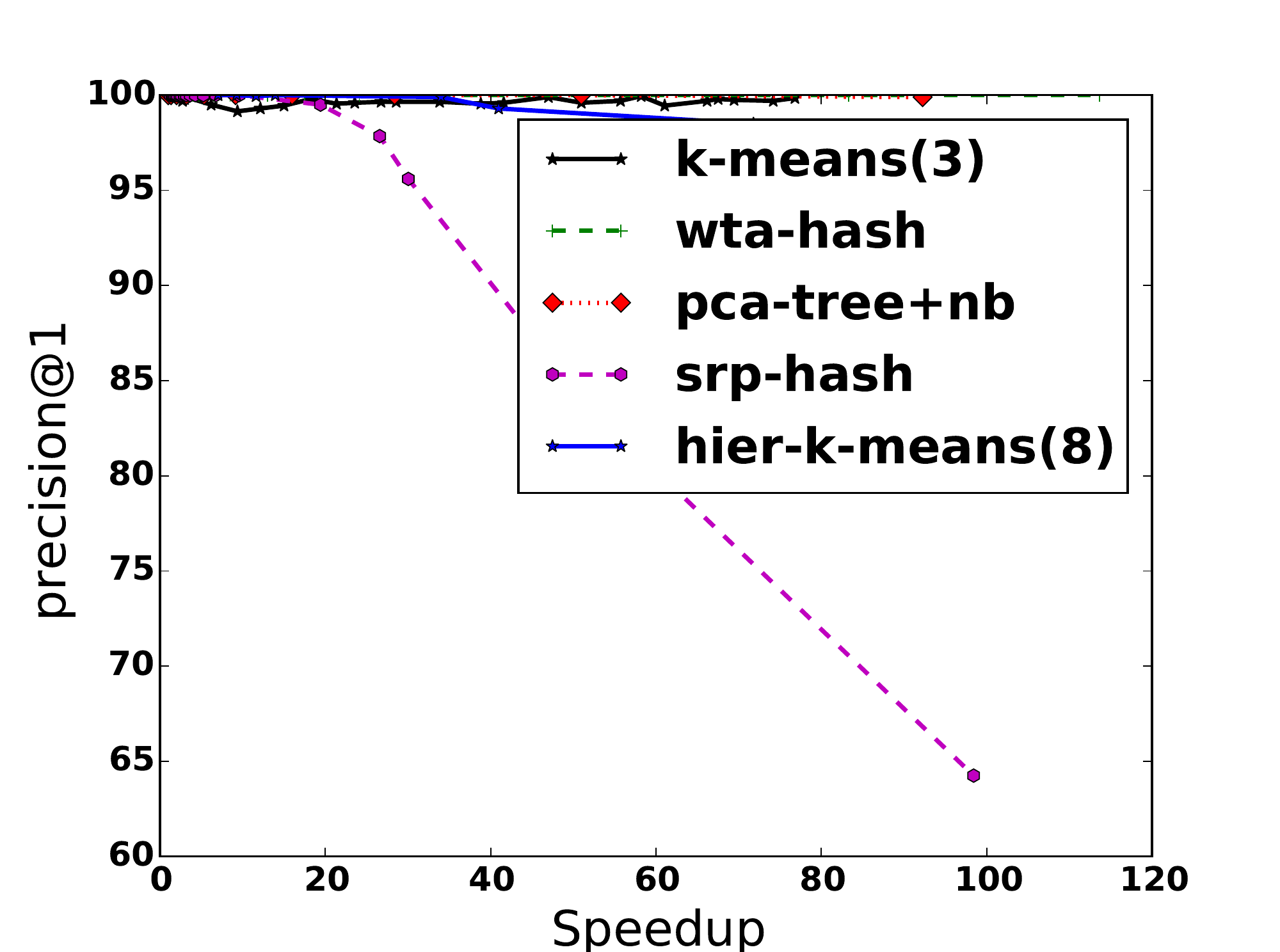}} 
\subfloat[]{\includegraphics[width = 1.8in]{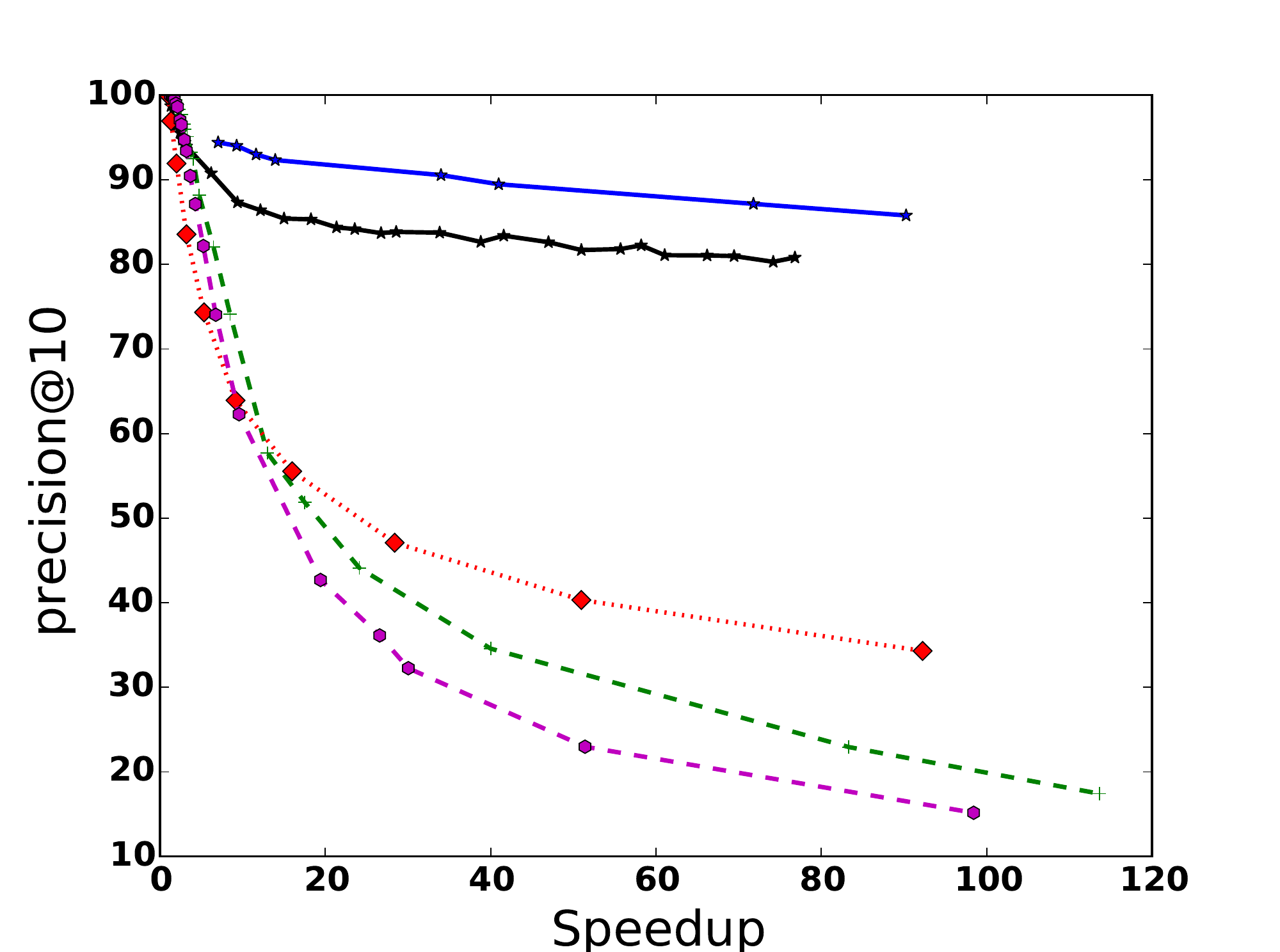}}   
\subfloat[]{\includegraphics[width = 1.8in]{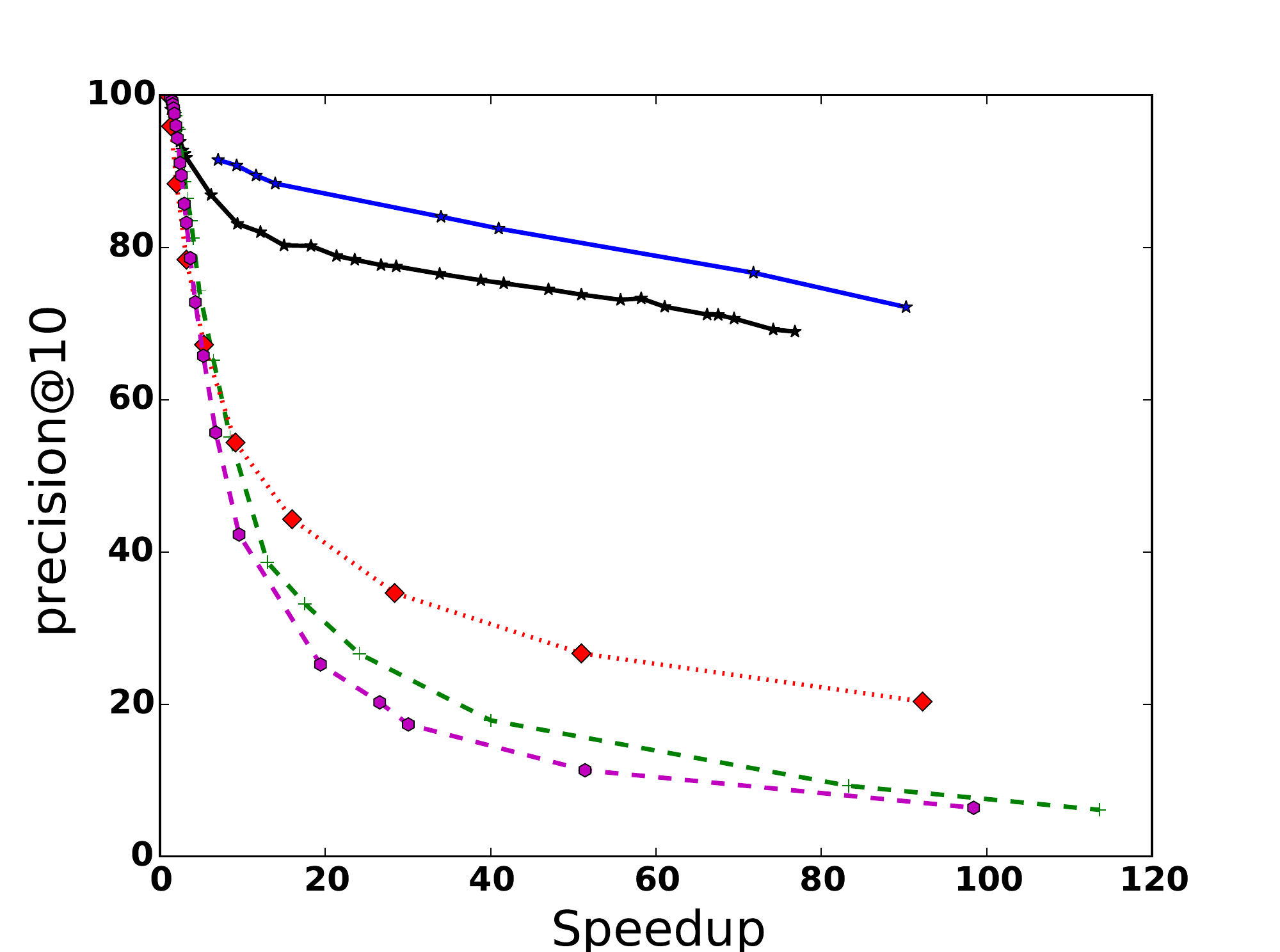}}\\
\caption{Speedup results in word embedding retrieval. (a-c) correspond to precision in top 1,10,100 MIPS respectively. $k$-means(3) means $k$-means algorithm that considers top 3 clusters as candidate set. and hier-$k$-means(8)s means a 2 level hierarchical k-means algorithm that considers top 8 clusters as candidate set.}
\label{ef2}
\end{figure}

In this experiment, we consider a word embedding retrieval task. As a first experiment, we consider using a query set of 2,000 embeddings, corresponding to a subset of a large database of pretrained embeddings. Note that while a query is thus present in the database, it is not guaranteed to correspond to the top-1 MIPS result. Also, we'll be interested in the top-10 and top-100 MIPS performance. Algorithms which perform better in top-10 and top-100 MIPS for queries which already belong to the database preserve the neighborhood of data points better. Figure \ref{ef2} shows the precision vs. speedup curve for top-1, top-10 and top-100 MIPS. From the results, we can see that data dependent algorithms ($k$-means and PCA-Tree) better preserve the neighborhood, compared to data independent algorithms (SRP-Hash, WTA-Hash), which is not surprising. However, $k$-means and hierarchical $k$-means performs significantly better than PCA-Tree in top-10 and top-100 MIPS suggesting that it is better than PCA-Tree in capturing the neighborhood. One reason might be that $k$-means has the global view of the vector at every step while PCA-Tree considers one dimension at a time.  

\begin{figure}[ht]
\subfloat[]{\includegraphics[width = 1.8in]{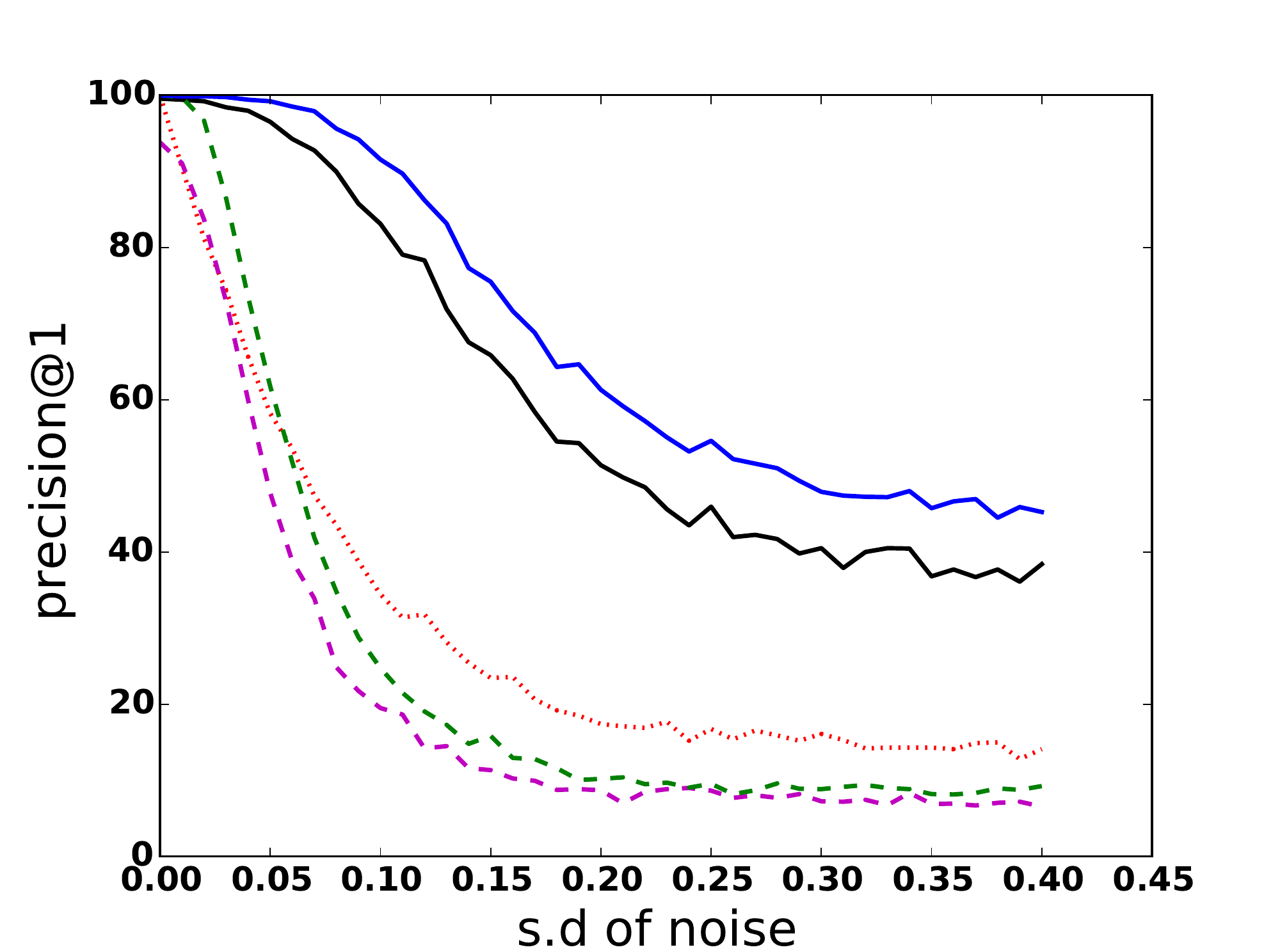}} 
\subfloat[]{\includegraphics[width = 1.8in]{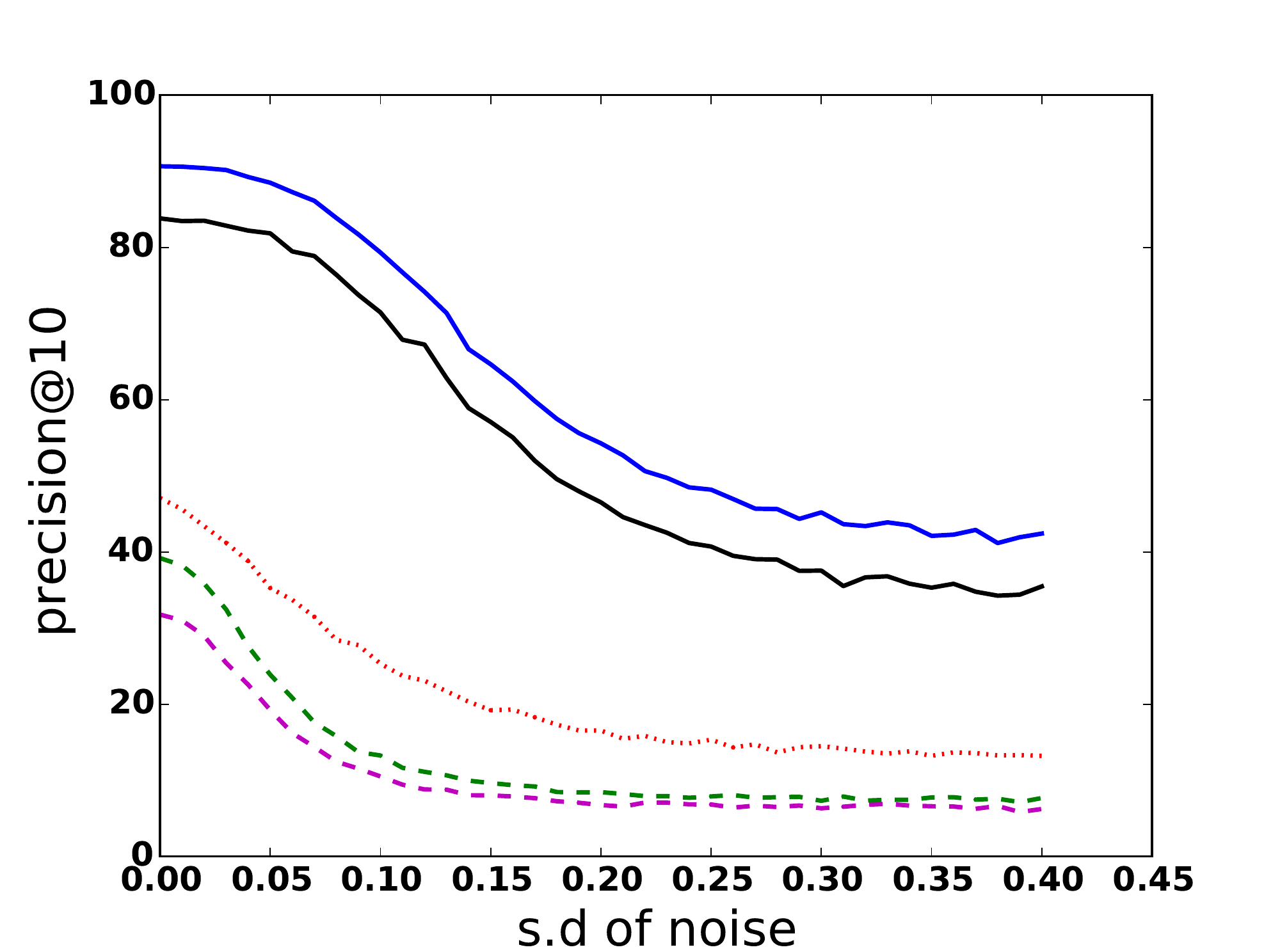}}   
\subfloat[]{\includegraphics[width = 1.8in]{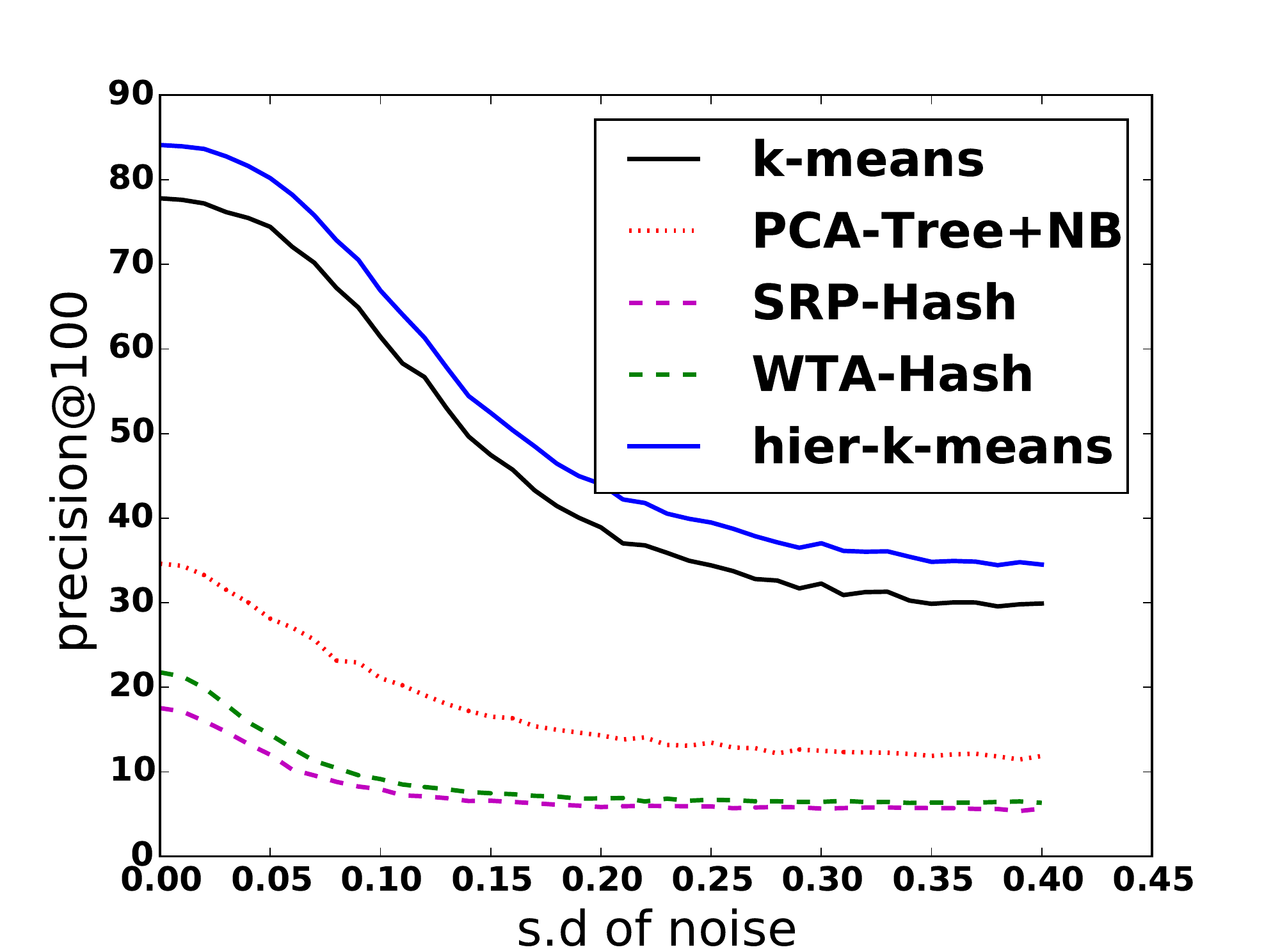}}\\
\caption{Precision in top-$K$ retrieval as the noise in the query increases. We increase the standard deviation of the Gaussian noise and we see that $k$-means performs better than other algorithms. }
\label{ef3}
\end{figure}

As the next experiment, we would like to study how different algorithms behave with respect to the noise in the query. For a fair comparison, we chose hyper-parameters for each model such that the speedup is the same (we set it to 30x) for all algorithms. 
We take 2,000 random word embeddings from the database and corrupt them random Gaussian noise. We vary the scale of the noise from 0 to 0.4 and plot the performance. Figure \ref{ef3} shows the performance of various algorithms on the top-1, top-10, top-100 MIPS problems, as the noise increases. We can see that $k$-means always performs better than other algorithms, even with increase in noise. Also, the performance of $k$-means remains reasonable, compared to other algorithms. These results suggest that our approach might be particularly appropriate in a scenario where word embeddings are simultaneously being trained, and are thus not fixed. In such a scenario, having a robust MIPS method would allow us to update the MIPS model less frequently.

\section{Conclusion and future work}
In this paper, we have proposed a new and efficient way of solving approximate $K$-MIPS based on a simple clustering strategy, and showed it can be a good alternative to the more popular LSH or tree-based techniques. We regard the simplicity of this approach as one of its strengths. Empirical results on three real-world datasets show that this simple approach clearly outperforms the other families of techniques. It achieves a larger speedup while maintaining precision, and is more robust to input corruption, an important property for generalization, as query test points are expected to not be exactly equal to training data points. Clustering MIPS generalizes better to related, but unseen data than the hashing approaches we evaluated.

In future work, we plan to research ways to adapt on-the-fly the clustering  for our approximate $K$-MIPS as its input representation evolves during the learning of a model, leverage efficient $K$-MIPS to speed up extreme classifier training and improve precision and speedup by combining multiple clusterings.

Finally, we mention that, while putting the final touches to this paper, another very recent and different MIPS approach, based on vector quantization, came to our knowledge \citep{quant}. We highlight that the first arXiv post of our work predates their work. Nevertheless, while we did not have time to empirically compare to this approach here, we hope to do so in future work.

\section*{Acknowledgements}

The authors would like to thank the developers of Theano \citep{bergstra+al:2010-scipy} for developing such a powerful
tool. We acknowledge the support of the following organizations for research funding and computing
support: Samsung, NSERC, Calcul Quebec, Compute Canada, the Canada Research Chairs and
CIFAR.

\bibliographystyle{plainnat}
\bibliography{refs}

\end{document}